\documentclass{svjour3}                     
\smartqed  

\usepackage{times}
\usepackage[hyphens]{url}
\usepackage{latexsym}
\usepackage{graphicx}
\usepackage{epsfig}
\usepackage{subfig}
\usepackage{multirow}
\usepackage{booktabs}
\usepackage{tabularx}
\usepackage[round]{natbib}
\usepackage{rotating}
\usepackage{verbatim}
\usepackage{CJK}
\usepackage{fontenc}
\usepackage{pinyin}

\usepackage{footnote}
\makesavenoteenv{tabular}

\begin{document}
\title{Creating a Live, Public Short Message Service Corpus: The NUS SMS Corpus\thanks{The authors gratefully acknowledge the support of the China-Singapore Institute of Digital Media's support of this work by the ``Co-training NLP systems and Language Learners'' grant R 252-002-372-490.}}

\author{Tao Chen \and Min-Yen Kan}

\institute{Tao Chen \at
              AS6 04-13
              Computing 1, 13 Computing Drive
              National University of Singapore
              Singapore 117417 \\
              Tel.: +65-83988018\\
              \email{taochen@comp.nus.edu.sg}
           \and
           Min-Yen Kan \at
              AS6 05-12
              Computing 1, 13 Computing Drive
              National University of Singapore
              Singapore 117417\\
              Tel.: +65-65161885\\
              Fax: +65-6779-4580\\
              \email{kanmy@comp.nus.edu.sg}
}

\date{Received: date / Accepted: date}
\maketitle

\begin{abstract}
Short Message Service (SMS) messages are largely sent directly from
one person to another from their mobile phones.  They represent a
means of personal communication that is an important communicative
artifact in our current digital era.  As most existing studies have
used private access to SMS corpora, comparative studies using the same
raw SMS data has not been possible up to now.

We describe our efforts to collect a public SMS corpus to address this
problem.  We use a battery of methodologies to collect the corpus,
paying particular attention to privacy issues to address contributors'
concerns.  Our live project collects new SMS message submissions,
checks their quality and adds the valid messages, releasing the
resultant corpus as XML and as SQL dumps, along with corpus
statistics, every month. We opportunistically collect as much metadata
about the messages and their sender as possible, so as to enable
different types of analyses.  To date, we have collected about 60,000
messages, focusing on English and Mandarin Chinese.

\end{abstract}


\section{Introduction}
\label{sec:introduction}
\subsection{SMS in communication}
Short Message Service (SMS) is a simple and ubiquitous modern form of world-wide communication.  As mobile handsets increasingly become cheaper to manufacture, and as the secondhand handset market distributes them more widely downstream, SMS emerges as the most widely-used form of digital communication next to voice telephony.  As of 2010, there were 5.3 billion active users of SMS, globally sending 6.1 trillion messages as estimated by the International Telecommunication Union (ITU)\footnote{\url{http://www.itu.int/ITU-D/ict/material/FactsFigures2010.pdf}}. In the same year, Americans alone sent 1.8 trillion messages \footnote{\url{http://www.ctia.org/media/press/body.cfm/prid/2021}}.
Even in China, 98.3\% of phone owners use SMS service, averaging 42.6
received messages weekly \footnote{\url{http://anti-spam.cn/pdf/sms1002.pdf}}.

The 160 character limit of SMS was designed by
Hildebrandt\footnote{\url{http://latimesblogs.latimes.com/technology/2009/05/invented-text-messaging.html}}
to accommodate the ``communication of most thoughts'' presumably in
English and other Latin languages. The limits as well as the early
difficulties with text input pushed users to be inventive and create
shortenings for communication.  The resulting genre of communication
has its own name: ``texting''~\citep{crystal2008txtng}, and has been
debated as to whether it is a friend or foe of language creation, use
and understanding.  With the advent of the smartphone, other informal
communications (e.g., tweets, instant messages), which were previously
computer mediated, now are enabled on mobile phones as well, and the
length restriction has been circumvented by standards that allow
multiple messages to be concatenated together. They share some
characteristics with SMS, such as the frequent use of emoticons,
unconventional short forms and misspelled words. Tweets, for instance,
have been dubbed as ``SMS of the
Internet''\footnote{\url{http://www.business-standard.com/india/news/swine-flu\s-tweet-tweet-causes-online-flutter/356604/}}.

SMS messages have been studied since the late 1990s, when SMS
communication became widely available among cellular carriers. The
ubiquitous, personal and informal nature of SMS attracted researchers'
attention, resulting in sociographic, linguistic and usability
research topics -- e.g., the impact of SMS on social culture,
linguistics analysis, text-entry improvement, name-entity extraction,
SMS normalization, authorship detection and spam messages detection.

With such a clear societal impact, however, SMS are underwhelming
studied.  It is clear why -- gathering a large corpus to study is
difficult and fraught with confidentiality problems, as messages are
often personal or contain confidential information.  Unsurprisingly,
there are thus few publicly available SMS corpora, discouraging
comparative studies. The unavailability of corpora also becomes an onus
on the aspiring researcher, forcing them to gather messages
themselves.

We now discuss three issues related to SMS corpus collection: 1) Why
are there so few public SMS corpora? 2) What factors make collecting
SMS so difficult? and 3) Can other corpora of communication vehicles,
such as tweets, replace SMS in studies?

SMS messages are primarily stored in mobile phone carriers' database;
only a small portion of them are stored in users' phones given limited
phone storage.  For obvious legal and privacy reasons, carriers cannot
release their SMS databases for research, as users' messages are often
personal, privileged and private and users would not want them known
to others. Even when SMS corpora are collected by researchers from
phone owners, such researchers have to maintain exclusive access for
the same private reasons. Thus, the private nature of SMS results in
the scarcity of public SMS corpora.

In this article, {\it SMS collection} in particular refers to
gathering messages directly from phone owners, largely for research
purposes. There are two major explanations for the difficulty of SMS
collection.  As researchers want the text and the metadata for study,
privacy is a major concern, as the majority of phone owners are
uncomfortable with having their private messages revealed, even when
the purpose is for research and academic pursuits.  Additionally, in
collecting SMS directly from users, the target messages to be
collected are stored on the users' mobile phones, which means that the
collection of a large corpus requires the cooperation of many phone
owners, and the support of software to ease the collection over many
potential mobile phone platforms.

In recent times, Twitter and social networks such as Facebook have
made the genre of the short message even more ubiquitous.  Tweets and
status updates are closely related to SMS, sharing the characteristic
terse style of messages.  However compared to SMS, tweets are
remarkable easier to gather since Twitter release API for accessing
data. So a natural question arises: can tweets replace SMS for related
studies?  Perhaps for some purposes, the answer can be ``yes'', but
they have fundamental differences which still preserves SMS as an
important medium to study.  First, SMS is a private communication
between two parties, which may contain very sensitive topics or
information (e.g., bank account, email address), hence its difficulty
for collection.  In contrast, tweets and a large portion of social
network messages and comments are decidedly broadcast media, and hence
far less private and sensitive. Second, though both SMS and tweets
have the 140 characters restriction, they still differs in
length. Ayman reported that the bulk of tweets are around 40
characters long\footnote{\url{http://www.ayman-naaman.net/2010/04/21/how-many-characters-do-you-tweet/}}
in a corpus of 1.5 million tweets, while \citet{Tagg2009} mentioned
that average words of SMS is 17.2 words in her corpus of 11,067
messages.  \citet{BachandGunnarsson2010} validates this, pointing out
that SMS messages were more likely to be very short -- containing only
one word -- compared to tweets, due to the more personal and
conversational aspects of SMS.  If the understanding of personal
informal communication and how it is evolving is important, then SMS
deserves to be collected and studied for the public good.

A public SMS corpus is needed to fill this gap for research
material. which will benefit all the researchers who are interested in
SMS studies. In 2004, our former project established a SMS collection
project for this aim, gathering and publishing a corpus of 10,117
English SMS, mainly from students in our university~\citep{How2005}.
The corpus was released freely online. It was the largest publicly
available English SMS corpus until 2010, and used in a number of SMS
studies. However, mobile technology has undergone a great development
in the past six years, especially with the advent of smartphone, which
have featured newer text input technologies and have influenced
people's texting habits. Moreover, there is an extreme scarcity of
Chinese SMS corpora in the public domain.

Considering this, we resurrected the SMS collection project in October
2010, reviving it as a live corpus collection project for both English
and Mandarin Chinese SMS. The ``live'' aspect of our project
emphasizes our aim to continually enlarge and publish the corpus for
end users.  Our SMS collection makes use of an array of collection
methodologies, leveraging current technology trends, with the aim of
making the resultant corpus be more representative of SMS worldwide
(rather than a specific community\footnote{In contrast, our 2004
  corpus was collected locally within the University in Singapore, not
  representative of general worldwide SMS use.}), to enable more
general studies.  Our current corpus also features improved collection
methodology, making the collection process more accurate with fewer
transcription errors, and is continuously released publicly with an
effective anonymization process.

As of October 2011, we have collected 28,724 English messages and
29,100 Chinese messages, resulting in the largest public SMS corpora
(to our knowledge), in terms of both English and Mandarin Chinese
languages, independently.

Our article reports on the contributions of our corpus and its
collection methods.  In particular, we:
\begin{itemize}
  \item use a battery of methodologies to collect the corpus, paying
    particular attention to privacy issues to address contributors'
    concerns (section~\ref{sec:methods});
  \item create a website to document and disseminate our gradual
    achievement, enabling direct, online browsing of the messages. We
    also release the resultant corpus as XML and as SQL dumps, along
    with salient corpus statistics, on a regular monthly schedule
    (section~\ref{sec:statistics}); and
  \item exploit a good Chinese crowdsourcing website for language
    data collection and compare it with its more well-known, U.S.
    counterpart (section~\ref{sec:discussion}).
\end{itemize}


\section{Related Work}
\label{sec:related}

\subsection{Comparison of SMS Corpora}

While the scope of related work to texting in general is vast, for the
purposes of this report, we limit the scope of our review to scholarly
publications concerning SMS studies.  This makes the task feasible and
allows us to further break the review down into particular facets of
interest.  In particular, we pay attention to the size and language of
each collection, the characteristics of its contributors, how the
respective authors have collected their corpus, and the respective
corpus' availability.  An important goal of this chapter is to provide
a comprehensive inventory of SMS corpora that have been collected and
pointers to their associated studies.

A survey of the literature shows that publicly-available SMS corpora
are scarce.  The resulting scarcity motivates subsequent researchers
to collect their own SMS corpora for their specific projects, but as
these projects are often one-off, the resulting collections are also
not publicly available, creating a vicious cycle.  We are interested
in the underlying facets of their contributors' identity and the
collection methods used.  For ease of reference, we have also compiled
and listed the existing corpora in Table~\ref{table:known corpora}
(publicly available corpora are indicated by an asterisk).

\begin{itemize}

\item The {\bf Size} of existing corpora is tiny when comparing to the
  Twitter corpus with 97 million of posts~\citep{Petrovic2010}. Even
  the largest SMS corpus consists of ``only'' 85,870
  messages~\citep{Liu2010}. 50\% of the corpora contain less than 1000
  messages, and only five corpora comprise of more than 10,000
  messages. We attribute the small scale of these corpora to the
  aforementioned difficulty of collecting SMS. However when the corpus
  is small, the resultant findings of the studies are often not
  statistically significant~\citep{durscheid2011}.

\item The {\bf Language} of the corpora ranges from European
  languages (English, French, Germany, Italian, Polish, Swedish, etc),
  Asian languages (Mandarin Chinese), to African ones (Kiswahili and
  Xhosa). However, European languages dominate, with only two Chinese
  corpora and two African corpora being the exceptions.  A corpus can
  be classified as monolingual and multilingual, describing how many
  languages are exhibited in its component messages. Most corpora are
  monolingual, and to our knowledge, only five existing corpora are
  multilingual~\citep{Deumert2008, Elvis2009, BachandGunnarsson2010,
    Barasa2010, Bodomo2010, durscheid2011}.

\item {\bf Contributors} of the SMS collections can be categorized
  into known and anonymous contributors. Families and friends are the
  most common known contributors~\citep{Segerstad2002, ZicFuchs2008,
    Gibbon2008, Tagg2009, Barasa2010}. Others include
  colleagues~\citep{Ju2009, BachandGunnarsson2010}, students in a
  specific university~\citep{Thurlow2003, How2005, Gibbon2008},
  recruited teenagers~\citep{Kasesniemi2002, Grinter2003}.

  Anonymous contributors are those that the researchers do not
  personally know or need to make direct contact with.  The methods to
  carry out the collection are also more varied in such cases than in
  known contributor cases. An example is the corpus collected
  by~\citet{Ling2005}, which involves 2,003 anonymous respondents via
  a telephone survey. Another example is the corpus collected
  by~\citet{Herring2009} whose participants are anonymous audience
  members of an Italian interactive television program.  Perhaps the
  most important instance is the distributed effort by the sms4science
  project, an international collaboration aiming to build an SMS
  corpus alongside corpus-based research on the resulting corpus. The
  sms4science subprojects have been carried out in nine countries,
  such as Belgium~\citep{Fairon2006},
  Switzerland~\citep{durscheid2011},
  France\footnote{\url{http://www.alpes4science.org}}, Greece, Spain
  and Italy.  Anonymous contributors were recruited from across
  the subproject's country or region.

  The difference between the known versus anonymous contributor
  corpora affects the corpora' representativeness.  Known contributors
  usually share similar demographic features, such as similar age
  (teenagers or college students), same career (colleagues), and/or
  same geographic location (living in a same city).  Hence, the
  corpora from known contributors may not be representative of the
  general landscape of SMS usage.  This characteristic may be
  perfectly acceptable or desired for the particular project for which
  the corpus is collected, since the project may be restricted in a
  particular purpose or study. For instance, \citet{Grinter2003}
  collected 477 messages from 10 teenagers for studying how messages
  have been incorporated into British teenagers' lives.
  Corpora from anonymous contributor projects, such as the sms4science
  project, are more broad and aim to satisfy general studies.  We note
  that both aims can be achieved in an anonymous collection process,
  as when suitable demographics are taken per message and the corpus
  is sufficiently large, an identifiable subset of the corpus can
  still serve for specialized studies.

\item {\bf Collection Methods.}  The most interesting facet of the
  studies for our purposes is how they collect their corpus.  We
  observed three primary methods to collect SMS. The simplest approach
  is to simply transcribe messages from the mobile phone, by typing
  them into a web based submission form~\citep{Segerstad2002,
    How2005}, into a word processing or other electronic
  document~\citep{Segerstad2002, Deumert2008, Tagg2009, Elizondo2011},
  or even the simple method of writing them down on
  paper~\citep{Kasesniemi2002,Grinter2003,Thurlow2003,Deumert2008,Bodomo2010}. Transcription
  can also happen later to facilitate collection speed --
  ~\citet{Lexander2011} took photos of messages stored on
  participant's phone, for later transcription by researchers.  A
  second method by exporting or uploading SMS messages via
  software. The corpus collected by ~\citet{Jonsson2010} is such an
  example. They implemented a phone widget for providing
  location-based SMS trend service and collecting SMS messages as
  well, since messages will be uploaded to the server when using the
  service.  \citet{Sotillo2010} and \citet{Walkowska2009} gathered
  messages by collecting SMS exported from contributors using mobile
  phone software suites such as Treo desktop and Microsoft's My
  Phone. The third class of methods is to have contributors forward
  messages to a collection number. Messages usually are forwarded to
  researcher's own mobile phone~\citep{Segerstad2002, Walkowska2009,
    Barasa2010}, which may incur cost for the contributors.  They are
  typically compensated for their cost, thus the large-scale
  collection can be costly. The studies done by \citep{Fairon2006, durscheid2011}, 
  were in collaboration with mobile phone operators, such that
  contributors could forward their messages to the operator-central
  number for free, further lowering the barrier for potential contributors.

  Aside from these common methods, we observed other one-off methods
  used in particular studies. \citet{Ogle2005} collected broadcasted
  SMS by subscribing to the SMS promotion of several nightclubs;
  \citet{Ling2005} asked respondents to read their SMS aloud during a
  telephone survey; \citet{Herring2009} downloaded the audience's SMS
  from an SMS archive of an interactive TV program; and finally,
  \citet{Choudhury2007} collected SMS from an online SMS backup
  website.

  Which method is best?  Each method has its own merits and drawbacks.
  If the scale needed is small, transcribing a few messages is easiest,
  needing virtually no effort in preparation and cost.  However for
  medium to large corpus projects, this methodology is not scalable,
  being time-consuming and prone to both transcription and deliberate
  correction errors, despite instructions to transcribe messages
  exactly as displayed.

  Exporting via software support preserves the originality of messages
  and in certain cases, gather valuable metadata (sender and
  receiver's telephone number and sending timestamp).  Exporting also
  enables batch submission, easily enabling a stream of valid and
  accurate messages to be collected.  It also encourages continuous
  SMS contribution, especially when the export-and-collection process
  can be automated and continuous.  The key drawback with this method is that
 it is tied to the phone model, and thus creates a selection bias in the
  possible contributors.  Exacerbating this problem is that some phone
  models do not even have (free) software to export SMSes.

  Forwarding messages is also effective to maintain the original
  message, but may be costly if the sending cost needs to be recouped
  by the researcher.  Forwarding may be easy for newly-sent messages
  as the collection number can be added as a second recipient.
  However, many phone models allow only forwarding single messages,
  such that forwarding lots of individual messages may be tedious.
  This discourages the collection of many messages from a single
  contributor.

\item {\bf Availability.}  In terms of availability, the existing
  corpora range from private, partially public, to completely
  public. As displayed in Table~\ref{table:known corpora}, most
  existing SMS corpora are private access.  Without a doubt privacy
  and non-disclosure issues are the underlying cause.  On one hand, it
  is the responsibility of researchers to protect the contributors'
  privacy and the easiest way to achieve the aim is by not making the
  corpus public~\citep{durscheid2011}. On the other hand, researchers
  may not be able to get the consent from contributors to release the
  corpus or be restricted by the rules of their university's IRB
  (Institutional Review Board)~\citep{Sotillo2010}.
  
  We define partially public corpora as corpora that are not immediately
  freely accessible, and require some type of initiative on the part
  of the scholar to secure.  The two partially public corpora are the
  Belgium Corpus~\citep{Fairon2006} and Swiss
  Corpus~\citep{durscheid2011}, collected by two sub-projects of
  sms4science. The former was distributed as a Microsoft Access
  database on a CD-ROM and is purchasable but restricted to bona fide
  research purposes only. The latter corpus is browsable online for
  researchers and students, but not downloadable as an entire corpus.
  Online browsing is very convenient for reading a few SMS without the
  need to parse the raw data, but makes it difficult to obtain all the
  SMS for serious corpus study.  We feel that this limits the
  potential studies that could employ the corpus.
    
  Completely public corpora are freely, immediately and wholly
  accessible. \citet{shortis2001} published 202 English SMS as a
  webpage\footnote{\url{http://www.demo.inty.net/app6.html}}. Although the
  corpus is not directly downloadable as a file, we still consider it
  as public as all of the messages are displayed on the single
  web page. Other public corpora were released as a file for freely
  downloading but vary in the file format. Both the German
  Corpus~\citep{Schlobinski2001}\footnote{\url{http://www.mediensprache.net/archiv/corpora/sms_os_h.pdf}} and HKU
Corpus~\citep{Bodomo2010}\footnote{\url{http://www0.hku.hk/linguist/research/bodomo/MPC/SMS_glossed.pdf}}
were released in Portable Document Format (.pdf),
while the IIT Corpus~\citep{Choudhury2007}
 \footnote{\url{http://www.cel.iitkgp.ernet.in/~monojit/sms.html}} and
 our aforementioned 2004 NUS SMS
 Corpus\footnote{\url{http://www.comp.nus.edu.sg/~rpnlpir/downloads/corpora/smsCorpus}}
 were released as text and XML files, respectively. The HKU Corpus is
 the only corpus containing about 140 Chinese language messages, but
 these mix English words, as may often be the case in Hong
 Kong. Strictly speaking, there is no pure, Mandarin Chinese SMS
 corpus in the public domain.

  Another public corpus is 9/11 pager messages released by Wikileaks
  in 2009\footnote{\url{http://mirror.wikileaks.info/wiki/911}} with
  over half million messages. The intercepts cover a 24 hours
  surrounding the September 11th, 2001 attacks, ranging from exchanges
  among office departments, to fault reporting of computers as the
  World Trade Center collapsed.  A few research studies have been
  conducted on this 9/11 corpus. \citet{Back2010} investigated the
  emotional timeline of the messages and analyzed the negative
  reaction to 9/11 terrorist attacks. \citet{Back2011} also used
  automatic algorithms and human judgment to identify 37,606 social
  messages from the original corpus, and rated the degree of anger in
  the timeline. Although 37,606 messages is quite a large corpus, most
  of the intercepts are limited to people's reaction to that terrorism
  event.  Such topically-focused pager messages cannot replace a
  general collection SMS messages, thus the corpus is not suitable for
  most SMS related studies.
  
 \item {\bf Release Time.} For the seven public SMS corpora mentioned
   above, all of them were released after the completion of data
   collection. The static release favors protection of contributors'
   privacy, since a global and thorough anonymization could be
   conducted~\citep{Fairon2006, durscheid2011}. A live release, in
   which the corpus is continually published and updated during the
   collection process, faces greater challenge and risk in
   anonymization.
\end{itemize}  

\noindent Due to the individual and private nature of SMS, the
resulting collected corpora also contain private details, which may
make it easy to discover the identity of the sender or recipient.  For
these reasons, such resulting corpora also cannot be made public.  As
mentioned, this creates a vicious cycle, erecting a barrier to SMS
research, making SMS seem less significant than it is for understanding
our current era of communication.  It is clear that a publicly
available, large-scale corpus of SMS could lower this barrier, and
make the study of SMS more widely available to scholars of all
disciplines.


\begin{savenotes}
\begin{table}
\caption{Existing Corpora. An asterisk (*) indicates that the corpus is publicly available.}
\label{table:known corpora}
\centering

    \begin{tabular}{|p{2.2cm}|r|p{1.5cm}|p{2.8cm}|p{2.7cm}|}
    \hline
    Researcher(s) 								& Size 		&	Language(s)			& Contributors 								& Collection Method(s) 	 					\\ \hline
    \citet{Pietrini2001} 					& 500			&	Italian					& 15 to 35 years old 					& {\it Not mentioned}										\\ \hline
    \citet{Schlobinski2001}$\ast$ & 1,500 	&	Germany					& Students  									& {\it Not mentioned} 									\\ \hline
  	\citet{shortis2001}$\ast$     & 202			&	English					& 1 student, his peers and family	& Transcription 					\\ \hline
    \citet{Segerstad2002} 				& 1,152		&	Swedish					& 4 paid and 16 volunteers & Transcription, Forwarding 					\\ \hline
    \citet{Kasesniemi2002}				& 7,800		&	Finnish					& Adolescents (13-18 years old)					& Transcription			 		\\ \hline
    \citet{Grinter2003}						& 477			&	English					& 10 teenagers (15-16 years old)					&Transcription 					\\ \hline
    \citet{Thurlow2003}						& 544			&	English					&	135 freshmen									&Transcription									\\ \hline
   	\citet{Ogle2005}						
& 97			&	English					&
Nightclubs						&Subscribe SMS promotion
of nightclubs   \\ \hline
   	\citet{Ling2005}							& 867			& Norwegian				& Randomly select 23\% of 2003 respondents & Transcription 		\\ \hline
   	\citet{How2005}$\ast$					& 10,117  &	English	       	&	166 university students	&Transcription												\\ \hline
   	\citet{Fairon2006}$\ast$	    & 30,000  & French        	&3,200 contributors		&Forwarding   																	\\ \hline
   	\citet{Choudhury2007}$\ast$   & 1,000		& English			&Anonymous users in treasuremytext\footnote{http://www.treasuremytext.com} & Search the SMS from the 																										website \\ \hline
   	\citet{rettie2007}						&	278			& English					& 32 contributors		& {\it Not mentioned}					\\ \hline
   	\citet{Ling2007}							&	191			& English					& 25 undergraduates & Transcription \\ \hline
   	\citet{ZicFuchs2008}					&	6,000		& Croation				&University students, family and friends	&{\it Not mentioned}\\ \hline
   	\citet{Gibbon2008}						&	292			&Polish						&University students and friends & {\it Not mentioned}  \\ \hline
   	\citet{Deumert2008}						&	312			&English, isiXhosa		&22 young adults	& Transcription, Forwarding \\ \hline
   	\citet{Hutchby2008}						&	1250		& English				&30 young professionals (20-35 years old) &	Transcription \\ \hline
    \citet{Walkowska2009}					&	1700		& Polish				&200 contributors			& Forwarding, Software  \\ \hline
    \citet{Herring2009}						&	1452		& Italian				&Audiences of a iTV program	& Online SMS archives		\\ \hline
    \citet{Tagg2009}							&	10,628	& English				&16 family and friends & Transcription	\\ \hline
    \citet{Elvis2009}							&	600			& English, French, etc.			&72 university students and lecturers & Forwarding \\ \hline
    \citet{Barasa2010}						&	2,730		&	English, Kiswahili, etc.			& 84 university students and 37 young professionals	&Forwarding \\ \hline
    \citet{BachandGunnarsson2010} &	3,152		& Swedish, English, etc.			& 11 contributors		& Software  \\ \hline
    
    \citet{Bodomo2010}$\ast$			&853			&English, Chinese 					&87 youngsters\footnote{The contributors and collection method are for the 487 messages collected in 2002; later, another 366 messages were collected from 2004-2006 without mentioning the contributors and collection methods.} & Transcription \\ \hline
    \citet{Liu2010}								&85,870		&Chinese			& Real volunteers			&{\it Not mentioned}\\ \hline
    \citet{Sotillo2010}						&6,629		&English			&59 participants			& Software			\\ \hline
    \citet{durscheid2011}$\ast$		&23,987		&Germany, French, etc				&2,627 volunteers			& Forwarding			\\ \hline
    \citet{Lexander2011}					&496			& French			&15 young people						&Transcription	 \\ \hline
    \citet{Elizondo2011}					&357			&English			&12 volunteers							&Transcription	 \\ \hline

    \end{tabular}
\end{table}
\end{savenotes}

\subsection{Crowdsourcing SMS collection}
From the above summary, we can see that a variety of approaches have
been employed to collect SMS for study.  In collecting any large-scale
corpora, it is necessary to distribute the task among a large group.
This is aptly illustrated by the sms4science project which involves
thousands of contributors.  As we aim to create an authoritative SMS
corpus to enable comparative studies, it is vital that the corpus also
be large.  Thus our methodology should also follow this distributive
paradigm.

Crowdsourcing, the strategy of distributing a task to a large
``crowd'' of human workers via some computer-mediated sources, has
emerged as a new driver of computation.  In tasks where raw compute
power cannot succeed, but where an aggregate of human judgments or
efforts can, crowdsourcing can be used.  It uses the computing medium
to connect many workers to a task necessitating human processing.  For
our particular instance of SMS collection, we can employ crowdsourcing
to connect many potential contributors of SMS to a collection
framework.

The term ``crowdsourcing'' actually subsumes several forms
~\citep{Quinn2009}, of which the Mechanized Labor form is most relevant
to our project.  This form is defined by its use of a (nominal)
monetary  to motivate distributed workers do their task. The most
notable example of mechanized labor in practice is embodied in Amazon
Mechanical Turk (hereafter,
MTurk)\footnote{\url{http://www.mturk.com}}, an online marketplace for
employers (a.k.a. requesters) to publish small tasks, and workers
(a.k.a. 
Turkers) to choose and complete the tasks.  MTurk has become
a popular crowdsourcing platform for its low cost and diverse
workforce.

Of late, MTurk has been employed for many uses within scholarly work.
We only focus on works concerning data collection, in particular, the
collection of language-related data. In 2010, a special workshop was
held with the North American Annual meeting of the Association of
Computational Linguistics (NAACL), entitled Creating Speech and
Language Data With Amazon's Mechanical Turk.
\citet{Callison-Burch2010} categorized the data collected in this
workshop into six types: Traditional NLP tasks, Speech and Vision,
Sentiment, Polarity and Bias, Information Retrieval, Information
Extraction, Machine Translation. Though strictly speaking, SMS
collection is not subsumed by any of the six types, the success
and variety of the corpora created in the workshop as well as in other
studies convince us that MTurk is a suitable platform for our project.



\section{Methodology}
\label{sec:methods}

We focus on collecting both English and Chinese SMS messages, for
expanding our 2004 English SMS corpus and addressing the extreme
scarcity of public Chinese SMS corpus.  Our aim is to create a SMS
corpus that is: 1) representative for general studies, 2) largely free
of transcription errors, 3) accessible to the general public without
cost, and 4) useful to serve as a reference dataset. Several
strategies were used in our collection process to achieve these goals.

First, we did not restrict our collection to any specific topics, to
encourage diversity among the messages. Also we did not limit to known contributors, but instead tried to diversify contributor backgrounds to fulfill the first aim of making contributor demographics similar to the general texting population. We used three different
technical methods to collect SMS: 1) simple transcription of an SMS into a collection web site, 2) exporting of SMS directly from
phone to a file for submission, and 3) uploading lists of SMS as an email
draft, for editing and eventual submission via email initiated by the
contributor.  The latter two collection strategies also favor the
collection of whole SMS streams during an interval,
favoring an unbiased collection of messages.  They also collect the messages
as-is from the phone's memory, minimizing the chance of
transcription or entry errors, satisfying the second aim.  To achieve
the third aim, we created a program (discussed below) to automatically
replace any identifiers and sensitive data with placeholders and to
encrypt identifiable metadata with each SMS.  With these minor
modifications, contributor's privacy issues are
mollified and allow us to release the corpus to the general public.
Finally, to ensure that the corpus satisfies our fourth aim of being a viable reference corpus, we
release archived, static versions of our continually-growing corpus on
a monthly basis.  In the following, we present these strategies in
more detail.

\subsection{SMS Collection Requirements}

We did not restrict contributors to send only SMS on certain
topics. This helps to keep the collected messages representative of actual
content~\citep{Barasa2010}, and diversify the corpus in
content. Moreover, we required contributors to fill out a demographic
survey about their background (e.g., age, gender, city, country),
texting habits (input method, number of SMS sent daily, years of using
SMS) and information about their phones (brand, model and whether it
is a smartphone or not). Such answers form a profile associated with
the bulk of the SMSes collected in our corpus, which we feel can
facilitate sociolinguistics studies.

We did, however, require that the submitted messages be personal, sent
messages only. The ``sent-message'' restriction is required for two
important reasons. Ethically speaking, the submission of received
messages is disallowed as the consent of the sender is not guaranteed,
and may violate the trust and rights of the original
sender.  Technically speaking, as we also aim to have as complete
demographics on the SMSes collected, we would also be unable to
contact the senders to have them complete the same demographic survey, which makes received messages even less appealing to collect.
The ``personal'' restriction means the messages are typed by the
contributors themselves and not of artificial or commercial nature, or
chain messages to be forwarded (e.g., blessings, jokes, quotes) that
may be available on the Internet.

\subsection{Source of Contributors}

We aim to create a corpus which reflects the general characteristics
of English and Chinese SMS messages, to be used for a wide variety of
studies.  Therefore, contributors are expected to have diverse
backgrounds, of a wide range of ages, and living in various geographic
location. As crowdsourcing methods pull from a variety of sources, we
deemed this strategy as suitable for SMS collection.

Probably the most well known crowdsourcing platform is Amazon's
Mechanical Turk (henceforth, MTurk), which allows users to publish
tasks and for the members of the general public to do the tasks,
usually for a nominal fee.  The use of MTurk as a crowdsourcing
technique has been widely documented in the computer science
literature. It is also an ideal place for conducting our English SMS
collection. We also published a few tasks in another mechanized labor
site, ShortTask\footnote{\url{http://www.shorttask.com}}, to diversify
the background of contributors. Considering its similarity to MTurk
and the limited usage in our current collection methods, we do not
discuss it further in this paper.

A demographic survey of MTurk workers (known colloquially as Turkers)
conducted by \citep{IpeirotisDemographics2010} reveals that the
respondents are from 66 countries with a wide distribution in age and
education levels, but that the majority of them are from
English-speaking countries (46.8\% American and 34.0\% Indian).
However, the study also suggests the scarcity of Chinese workers,
which has also been validated by researchers --
\citep{Gao2010,Resnik2010} who have pointed out that there are few
Chinese speaking Turkers and thus difficult to recruit. We also
performed a pilot study in MTurk, publishing two batches of tasks to
collect Chinese SMS messages, but we received few submissions,
validating the earlier reports of shortage of Chinese workers.  So
while MTurk is a good platform for collecting English SMS, and we have
to find a more suitable platform for gathering Chinese SMS.

In China, the same crowdsourcing form of mechanized labor goes by the
name of ``witkey'' \begin{CJK}{UTF8}{gbsn}(威客, {\it \Wei1 \Ke4} in
  pinyin) \end{CJK}, short for ``key of wisdom'', described as using
the wisdom of the masses to solve problems.  Among such Chinese
websites, Zhubajie\footnote{\url{http://www.zhubajie.com}} currently
stands out for its dominant market share (over 50\%) and huge
workforce (over 50 million)\footnote{According to the China Witkey
  Industrial White Paper conducted by iResearch
  \url{http://www.iresearch.com.cn}}.  Zhabajie also categorizes tasks
within its own ontology, and one specific category relates to SMS (more details
in section~\ref{sec:discussion}). Therefore, we chose Zhubajie as the crowdsourcing platform for collecting Chinese
SMS.

Besides anonymous workers in MTurk, ShortTask and Zhubajie, we also exploit our
local potential contributors in Singapore.  English and Chinese are
two of the official languages of Singapore, making it an ideal place
to recruit contributors for our corpus collection. We recruited contributors by
emailing students in our department.  They were requested to submit
either English or Chinese SMS.  Participants from all three above
sources were reimbursed a small sum of money for their contributions.

Finally, we also wanted to explore whether people would be willing to
contribute SMSes purely for the sake of science (without
remuneration).  To test this, we sent email invitations to Internet
communities of linguists and mobile phone owners.  These communities
comprised of the well-known {\it
  corpora-list}\footnote{Corpora@uib.no} (an international mailing
list on text corpora for research and commercial study), {\it
  corpus4u} (a Chinese corpora forum), {\it
  52nlp}\footnote{\url{http://www.corpus4u.org}} (a Chinese
collaborative blog in natural language
processing\footnote{\url{http://www.52nlp.cn}}), and two Chinese
popular mobile phone forums -- {\it
  hiapk}\footnote{\url{http://bbs.hiapk.com}} and {\it
  gfan} \footnote{\url{http://bbs.gfan.com}}.

\subsection{Technical methods}
Our collection methods can be categorized into three separate methods
of collection.  We want our methods to be simple and convenient for
the potential contributor and allow us to collect SMS accurately
without transcription errors.

\begin{itemize}

\item {\bf Web-based Transcription.} The simplest collection method is
  transcribing messages from phone.  We designed a web page for
  contributors to input their messages. Contributors were asked to
  preserve the original spelling, spaces and omissions of SMS, and
  standardized emoticons by following a transcription code table of
  our design (e.g., any simple smiling emoticon should be rendered as
  ``:)'').

  As it is simple to implement, we adopted this transcription method
  as the collection method we used in our pilot, when we restarted our
  collection in 2010.  We published a series of small tasks in MTurk
  to collect Chinese and English SMS, to test the waters and refine
  problems with our instruction set.  However, when reviewing the
  submitted messages before Turker payment, we found a serious
  problem: a high rate of cheating.  Some of the English Turkers had
  just typed messages like blessings, jokes, friendship, quote that
  were verbatim copies of ones that were publicly available in some
  SMS websites.  These messages would not be representative of
  personally-sent messages, as required in the task's documentation.
  For the Chinese SMS, it was apparent that some English speakers
  pretended qualify as Chinese speakers, copying Chinese sentences
  from Internet websites or completing the task without actually
  submitting any content.  For both languages, we spent a non-trivial
  amount manpower (about 3.5 hours for 70 English submissions, and half an hour for 29 Chinese
  submissions) to inspect such messages, to validate the
  submissions as original, checking against identical or very similar
  messages publicly available the web.  For our pilot tasks, our
  rejection rate of English and Chinese messages was 42.9\% and 31.0\%,
  respectively -- clearly suboptimal considering the time and effort
  needed to review submitted SMSes and the potential ill-will
  generated with Turkers who performed our tasks but whom we deemed as
  cheating.

  A final problem is that transcription is prone to typos and
  deliberate corrections, and discourages contributors from inputting
  a lot of messages, since the re-typing is tedious. From these pilot
  tasks, we learned that we needed better collection methods that
  ensured message accuracy and demanded less validation.

\item {\bf SMS Export.} With supporting software, some mobile devices
  can export received or sent SMS as files in various formats such as
  TXT, CSV, XML, HTML.  This capability allows us to define this
  second collection method, SMS Exporting.  It involves two
  steps. First, contributors export SMS from phone as a readable
  archive, and optionally, censor and delete SMS that they do not want
  to contribute. Second, contributors upload the archive and take a
  web-based demographic survey.  We recruited contributors via both
  crowdsourcing websites and by regular, email invitations.
  Our unified (for both emailed invitations and the crowdsourcing
  tasks) description asked contributors to participate if they can
  export messages as a readable file (e.g., CSV, XLS).  While such
  exporting capabilities are becoming more prevalent, not all phone
  models have such software.  Even when available, it is not always
  free nor easy to use.  Noting these difficulties, we thus prepared
  notes to ease contributors' potential difficulties for popular
  platforms.

  Demographics from our web-based transcription task fielded in MTurk
  shows that 60\% of English SMS workers and 47\% of Chinese SMS
  workers were Nokia owners. This phenomenon is in accord with Nokia's
  large market share and penetration in China and India (74\% of English 
  SMS workers in the pilot task are from India). Fortunately, Nokia
  provides the Nokia PC Suite\footnote{Currently being replaced by Nokia Suite
    \url{http://www.comms.ovi.com/m/p/ovi/suite/English}}, a free software package for SMS export
  and backup via computer, which works on most Nokia models and
  meets our requirements. In the task description, we therefore linked
  to the download site for Nokia PC Suite and we offered a
  webpage-based tutorial on how to export SMS using the software.

  Besides the advantage of high accuracy, and ease of batch
  submissions of SMS as mentioned in Section~\ref{sec:related}, SMS Export greatly
  helps us in validation. Since the archive has a specified format --
  which includes the telephone numbers of the sender and receiver, the
  send timestamp of the message -- it significantly lowers the barrier
  for submitting valid data and significantly raises the barrier for
  submitting false data.  For this reason, we expend significantly less
  effort in validating SMS submitted by this process.
  
\item {\bf SMS Upload.} With the growing popularity of smartphones,
  which have added functionality, we felt it would be a good idea to
  implement mobile applications (``apps'') that can contribute SMS
  directly.  At the current juncture, we have implemented an app for
  the Android platform.  Inspired by another app, SMS Backup, which is
  an open-source Android app for backing up SMS to
  Gmail\footnote{\url{http://mail.google.com}}, we adapted the code to
  create a new app, which we called {\it SMS Collection for Corpus},
  as a pilot software for smartphones.  We have released it as a free
  application in Google's Android
  Market\footnote{\url{https://market.android.com/search?q=pname:edu.nus.sms.collection}}.
  Figure~\ref{fig:app_en} shows a snapshot of our app.

  {\it SMS Collection for Corpus} works by uploading sent SMSes from
  the Android device to the user's Gmail\footnote{Hence a Gmail
    account is a prerequisite to this collection method.} as a draft
  email.  To allow the user to censor and delete message that she does
  not deem suitable for the corpus, the app purposely does not send
  the messages directly to our collection web site.  We do not receive
  the SMSes for the corpus until contributors act to send out the
  draft email to us.  The app also automatically anonymizes the
  metadata (telephone numbers) and replaces sensitive identifiers with
  placeholders.  The details of the anonymization process is described
  later in this section.

  As with the SMS Export collection method, it reduces the possibility
  of cheating and preserves the originality of messages. One advantage
  over SMS Export is its convenience; as there is no need to connect
  to a computer. Most importantly, its automatic anonymization makes
  contributors more confident about our strategy in privacy
  protection. For SMS collected in the other two methods, we
  employ a similar anonymization process, but after receiving the
  original SMS; in this method the anonymization procedure is run on
  the client smartphone, even before it reaches our collection server.

  To the best of our knowledge, {\it SMS Collection for Corpus} was
  the first application designed to collect a corpus from original
  sources. It is also easy to adapt the software to support
  internationalization, so that the user interface can support new
  languages for potential submitters in other languages.  In July
  2011, we did exactly this, we extended the UI (user interface) to support prompts in
  Dutch, to facilitate the
  SoNaR\footnote{\url{http://www.sonarproject.nl}} project language
  corpus project, to assist them to gather SMS in Netherlands.

\begin{figure}[h]
  \centering
    \includegraphics[width=0.4\textwidth]{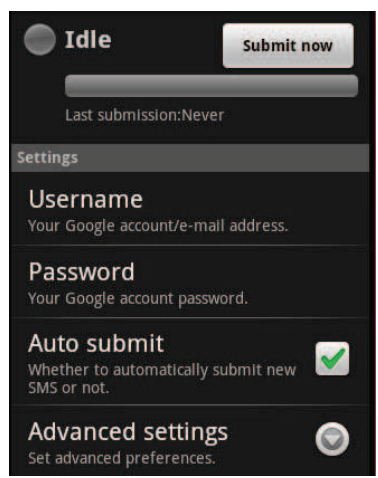}
     \caption{A screen capture of the {\it SMS Collection for Corpus}
       mobile application.}
    \label{fig:app_en}
\end{figure}

\end{itemize}

\subsection{Anonymization}

Since SMS often contains both personal and confidential information,
such as telephone numbers and email addresses, we need to anonymize
these data when included in submitted SMS.  We need to do so
systematically in order to guarantee privacy when publishing the
corpus, as we want to make the messages publicly available without
restriction on access or use.

For messages collected by SMS Export and SMS Upload, the SMS metadata
-- telephone number of sender and receiver, and send time -- is also
collected.  However, for the sender and recipient SMS metadata, we
need to replace the original data with a unique identifier for each
phone number so that privacy can be maintained, while preserving the
fact that multiple messages linked to the original, same source are
attributed correctly in the anonymized versions.  To solve this
problem, we adopt DES encryption to create a one-way enciphering of
the phone numbers, which replace the originals in the corpus.

For the SMS message body, sensitive data include dates, times, decimal
amounts, and numbers with more than one digit (telephone numbers, bank
accounts, street numbers, etc.), email addresses, URLs, and IP
addresses.  Privacy for these types of data is paramount, and as such,
we adopt a stricter standard in dealing with sensitive data in the
message itself.  Such information is programmatically captured using
regular expressions and replaced by the corresponding semantic
placeholders, as shown in Table~\ref{table:replacement code}. For
example, any detected email address will be replaced by the code
$\langle EMAIL \rangle$.  This process gives a level of protection
against publishing sensitive data.  While we remove such confidential
information, in general it is impossible to guarantee a contributor's
privacy with only a simple set of textual regular expressions.  In
particular, as person names are varied, language-specific and often
confusable with common words, we do not try to remove or replace
personal names.

All contributors were informed about the intended publishing of the
resultant corpus, its public availability and the above anonymization
procedure.  This process also aided our internal review board
application for exemption, as it was deemed that through this method,
that our collection procedure did not collect personally identifiable
information and was granted exemption from full review.  However, the
contributors may still not be entirely clear about the automatic
anonymization procedure after reading the description. To eliminate
their uncertainty and skepticism, we need a straightforward and
compelling way to show the anonymization in action. As we mentioned
before, our Android app integrates the anonymization process
internally, so potential submissions can be previewed as a draft email
before sending the SMS batch to the collection server. This manner
allows the actual collection data to be previewed, and more likely to
convince the contributor of the veracity of the project and collection
process.

Inspired by this, we created and deployed the website for the corpus
project in January 2011, at the very beginning of our data collection
process.  The website allows users to browse the current version of
the corpus.  In calls to contributors, we also link to this live
website so that potential contributors can view the scope, volume and
content of the current corpus.  We feel this is a strong factor in
both lowering the anxiety and reluctance of potential submitters and
raising awareness of the corpus in general.

\begin{table}[h]
\centering
 \caption{Replacement Codes}
 \label{table:replacement code}
\begin{tabular}{l*{3}{c}r}
\hline
Original Content     			&Example			 		&  	Replaced Code \\
\hline
Email Address 					&name@gmail.com				& 	$\langle EMAIL \rangle$ \\
URL            					&http://www.google.com		&  	$\langle URL \rangle$ \\
IP Address          			&127.0.0.1 					& 	$\langle IP \rangle$  \\
Time     						&12:30						& 	$\langle TIME \rangle$ \\
Date     						&19/01/2011					& 	$\langle DATE \rangle$ \\
Decimal 						&21.3						&  	$\langle DECIMAL \rangle$ \\
Integer over 1 Digit Long 		&4000 						&	$\langle \# \rangle$ \\
Hyphen-Delimited Number  		&12-4234-212				&	$\langle \# \rangle$ \\
Alphanumeric Number		 		&U2003322X					&	$U\langle \# \rangle$X \\
\hline
\end{tabular}
\end{table}

\subsection{Live Corpus}

A few words about the notion of a live corpus.
We feel that a live corpus is a emerging concept, in which the corpus grows, is 
maintained and released on regular, short intervals.  A truly live corpus connotes that
as soon as a new text is created, it becomes part of the distributed corpus. 
Such an interpretation can cause replicability problems, as different
researchers may use different versions of corpus.  Due to this problem, we have
chosen to release a new version of the corpus on a regular, monthly basis. This
strategy of having regular updates promotes interested parties to stay up to
date with the corpus developments, while allowing the easy identification of a
particular version, for papers that wish to use the corpus for comparative
benchmarking.  The release cycle further helps to demonstrate the trend of our
gradual achievement in SMS collection, which, in turn, also may spur more
contributors to help in our project.  It also allows us to batch corpus
administrative duties, such as proofchecking submitted SMS and re-computing
demographic statistics, which we describe later.



\section{Properties and Statistics}
\label{sec:statistics}

Given the variety of methods in our current corpus collection, and its
previous history within another former project within our group, it is
worthwhile to describe some properties of the resultant SMS collection
thus far.

The original corpus, as collected and documented in \cite{How2005},
was collected by an honors year project student, Yijue How, over the
course of her project from the fall of 2003 to the spring of 2004.
The main collection method was by SMS transcription to a
publicly-accessible website, largely from volunteers that authors
contacted directly, hence most volunteers were Singaporeans in the
young adult age range.  A few contributors gave a large set of
messages directly from the phones, foregoing the transcription
process.  This led to a distinctly bimodal message distribution of
having a small ``head'' of few contributors that submitted many
messages that represented depth in the collection, as well as a long
``tail'' of many contributors that submitted less than 100 messages
(The transcription website allowed the submission of 10 messages at a
go).  Each contributor was assigned an identifier such that the number
of messages by each contributor could be tracked.  Further details
about the demographics of the collection are available from How's
complete thesis \citep{How2004}.

We embarked on our current SMS collection project in October 2010.  At
the time of writing (October 2011), we have collected almost 60,000
messages, with the number still growing.  In this remainder of
section, we give statistics on the current corpus, the
demographics of contributors, furnish a cost comparison of the major
three sources we used and wrap up with a discussion on matters related
to the corpus release.

\subsection{Specification}
\begin{figure}[h]
  \centering
    \includegraphics[width=0.8\textwidth]{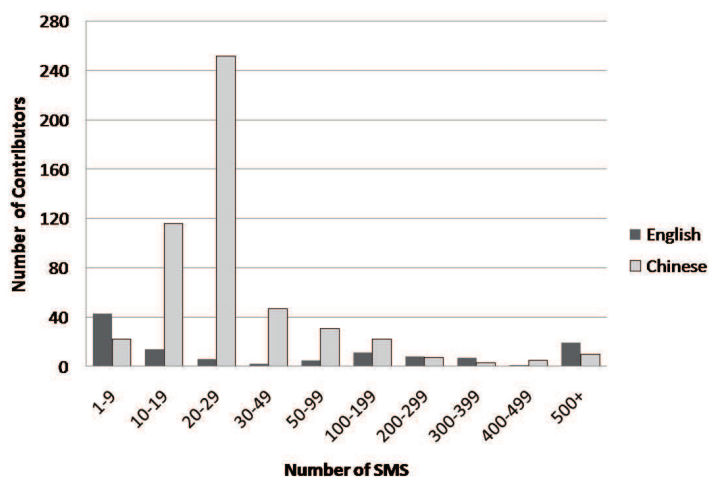}
     \caption{Contributors and SMS Distribution}
    \label{fig:distribution}
\end{figure}

As of October 2011, our corpus contains 28,724 English messages and 29,100  
Chinese messages. In total, 116 contributors contributed English SMS and on
average, each individual submitted 247.6 messages.  In comparison, the total
number of Chinese contributors is 515, with an average contribution rate 56.5
messages per person. Detailed histograms showing the number of messages
contributed per contributor are given in Figure~\ref{fig:distribution}.  
Similar to the previous project, both histograms show a peak at the very
left side, meaning that only a small proportion of people contributed the
bulk of the messages -- ``the short head''. Specifically, 54.3\% of English
contributors submitted less than 30 messages. This figure is 75.7\% for Chinese
contributors, which explains why the per contributor Chinese SMS figure is much less
than its English counterpart. 

The cause of this difference is related to our collection methods. As discussed 
previously, due to its simplicity, Web-based Transcription is an effective way
to obtain mass participation but makes it difficult to collect large amounts of
SMS from a single contributor, while both the SMS Export and SMS Upload methods
have the opposite characteristic.
We fixed the number of SMS per submission in our Web-based Transcription method 
to 2 or 5 English messages, and 10 or 20 Chinese messages. A small number was used
in the initial experiment for exploring a good ratio between monetary reward and
workload (number of messages). Using MTurk,
we published two batches of tasks to recruit 40 workers to follow our Web-based
Transcription to collect English SMS.  Unfortunately, due to the resulting high
level of cheating and effort expended in verification, we felt the utility of
this method was not tenable, so we stopped using this collection method for English
SMS.  In contrast, the Web-based Transcription was much more effective in
Zhubajie, perhaps due to the unavailability of sources to cheat on the task.  Up
to now, we have retained the use of Web-based Transcription in Zhubajie for the
resulting high-quality SMS.  This leads to the ascribed difference in
demographics in recruiting more Chinese contributors with a resulting smaller
per capita figure.

Table~\ref{table:sms and methods} demonstrates the number of messages collected 
by each of our methods. We see that 98.3\% (28,244) of English messages and 45.9\% (13,347) of
Chinese messages were collected by SMS Export and SMS Upload methods, which are
free of typos and contain metadata (ownerships of sender and receiver, and the
timestamp of when it was sent).  Table~\ref{table:sms and sources} then shows
the number of SMS and contributors per source: for the English SMS 39.3\% (11,275) were
from workers in MTurk, 58.1\% (16,701) were from local contributors, 1.0\% (280) were from
workers in ShortTask and the remaining 1.6\% (468) were from the Internet community.
For the Chinese SMS, workers from Zhubajie contributed 81.7\% (23,789) of the
SMS, local and Internet contributors submitted 12.2\% (3,544) and 5.9\% (1,712), respectively. 
Currently, only 55 Chinese SMS were contributed by users from MTurk, about
0.2\%. 

\begin{table}[h]
\centering
 \caption{Number of SMS by Collection Method}
 \label{table:sms and methods}
\begin{tabular}{l*{6}{c}r}
\hline
Method     		&  English SMS  &		Chinese SMS \\
\hline
Web-based Transcription &  480 				& 15,753  \\
SMS Export            	&  11,104 			& 12,344 	\\
SMS Upload          	&  17,140 			& 1,003  	
\\
\hline
\end{tabular}
\end{table}

\begin{table}[h]
\centering
 \caption{Number of SMS and Contributors by Source}
 \label{table:sms and sources}
\begin{tabular}{l*{6}{c}r}
\hline
Source     	&  English SMS  	& English Contributors 	& Chinese SMS &
Chinese Contributors\\ 
\hline
MTurk 						& 11,275 			
&75										
& 55  				 &19\\
ShortTask						& 280						&17											& 0						&0\\		
Zhubajie            & 0 						&0											& 23,789 			 &483	\\
Local          			& 16,701				&20											&	3,544  			 &10\\					
Internet Community 	& 468 					&4											& 1,712  	     &3\\
Total								& 28,724				&116											&29,100				 &515\\
\hline
\end{tabular}
\end{table}

\subsection{Demographics} 

Aside from submitting SMS, all contributors were required to fill out an online 
demographic survey about their background (e.g., age, gender, country, native
speaker or not, etc.), texting habits (input method, number of SMS sent daily,
years of experience using SMS) and their phone (brand, model, smartphone or
not).  Such answers form a user profile for the contributor which is linked to
each of their contributed SMS.  We accept and report the data for the
demographic survey as-is, assuming that contributors answered the questions to the
best of their knowledge.  

99.5\% of English messages and 93.8\% of Chinese messages thus have associated 
user profiles. The incompleteness arises from the separation of submitting SMS
and filling out the survey in the two collection modes of SMS Export and SMS
Upload. Some contributors submitted the messages but later did not do the
survey. The phenomenon was more prevalent in the Chinese Zhubajie. During our
initial usage pilots of Zhubajie, we approved contributors' SMS immediately, and
trusted them to do the survey later on.  To stem this problem, we later changed
our protocol to only approve the task after receiving both SMSes and the
demographic survey.  We also had updated the survey once, adding some questions to reveal  more detail
on some aspects.  The user profiles formed from the first version of the survey
thus lacks answers to a few questions.  To make all sets of the demographic data
comparable, we inserted ``unknown'' values to these missing questions as
well as to questions that were skipped by contributors.  

While it is not possible to conclusively say anything about the
demographics of SMS senders in general, our demographic survey allows
us limited insight into the characteristics of our corpus'
contributors.  The confounding factor is that our contributors come
largely from crowdsourcing sources, so both the self-selection of
participating in crowdsourcing and of SMS use contribute to the
demographic patterns we discuss below.

\begin{figure}[h]
  \centering
    \includegraphics[width=0.8\textwidth]{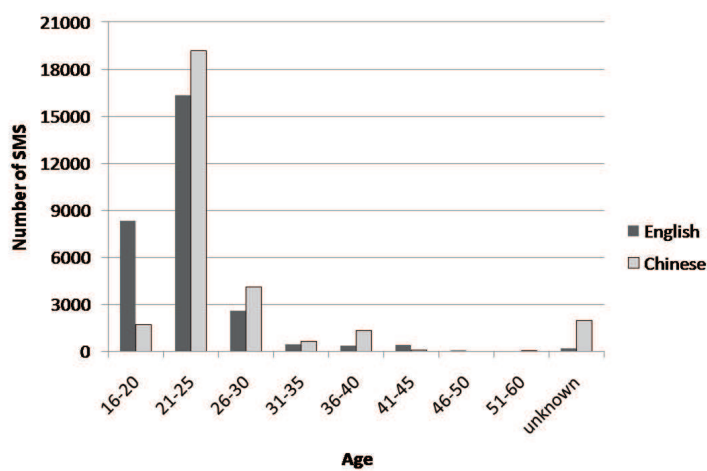}
     \caption{Age Distribution by SMS}
    \label{fig:age}
\end{figure}

We report both the country of origin, gender and age of contributors,
subdivided by the English or Chinese portion of the current corpus.
Our English SMS contributors are from 15 countries (in decreasing order of number of contributors): India, USA, Singapore, UK, 
Pakistan, Bangladesh, Malaysia, China, Sri Lanka, Canada, France, Serbia, Spain,
Macedonia and Philippines.
64.7\% of them are English native speakers. The top three countries are
Singapore (46.9\% of SMS), India (28.2\%) and USA (13.5\%). Our Chinese SMS
contributors are from 4 countries: China, Singapore, Malaysia and
Taiwan. However, the messages are overwhelming from China: China mainlanders
contributed 99\% of the messages, resulting in 99.8\% of messages originating
from native speakers.
As for gender, for the English portion of the corpus, 16.1\% come from females, 
71.1\% from males, and the remaining 12.8\% is unknown. For the Chinese portion,
34.3\% come from females, 59.4\% come from males, and 6.3\% is unknown. The age
distribution shows that the majority of contributors in both portions of the corpus are young adults, as displayed in
Figure~\ref{fig:age}. In particular, contributors aged at 21--25, taking up 39.7\%
of English and 56.7\% of Chinese SMS contributors, respectively,  make the 
biggest contribution to our corpus, submitting 56.9\% of English and
66.0\% of Chinese SMS respectively.  This may be reflective of the demographics
of crowdsourcing workers in both MTurk and Zhubajie 
\citep{IpeirotisDemographics2010,iResearch2010}.  There is an even higher skew
towards the 16--20 age group among English contributors than the Chinese
counterpart, largely due to the fact that 58.1\% English SMS originate from
local contributors, namely the students in our university.

\begin{figure}[h]
  \centering
    \includegraphics[width=0.8\textwidth]{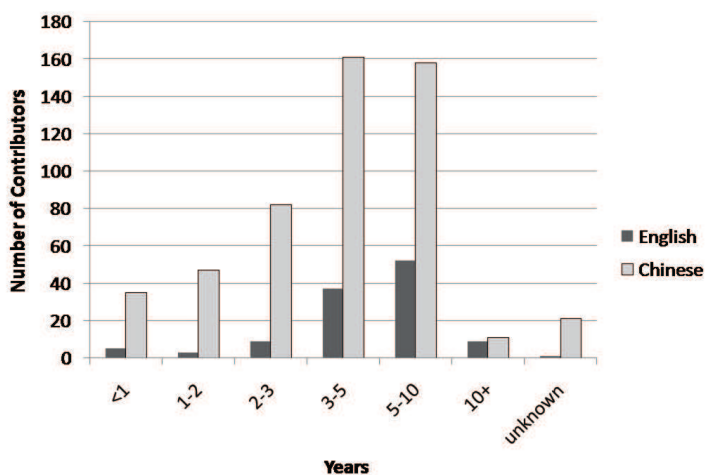}
     \caption{Contributors' Experience in Using SMS}
    \label{fig:experience}
\end{figure}

Our survey also reveals other pertinent details on the texting habits and phone models of contributors.  
We display the distribution of contributors' years of using SMS in
Figure~\ref{fig:experience}. The largest English SMS portion lies in
5--10 years (44.8\% of contributors), and the second largest portion is
3--5 years (31.9\%).  Similarly, most Chinese contributors have used
SMS for 3--5 years (31.3\%) and 5--10 years (30.7\%). This phenomenon is
in accord with the age distribution; that young adults represent the
majority of the contributors.  We may also posit that more of our
Chinese contributors have acquired their SMS--capable phone more
recently than our English contributors, as we see a smaller proportion of users in the 5--10 year range.

\begin{figure}[h]
  \centering
    \includegraphics[width=0.8\textwidth]{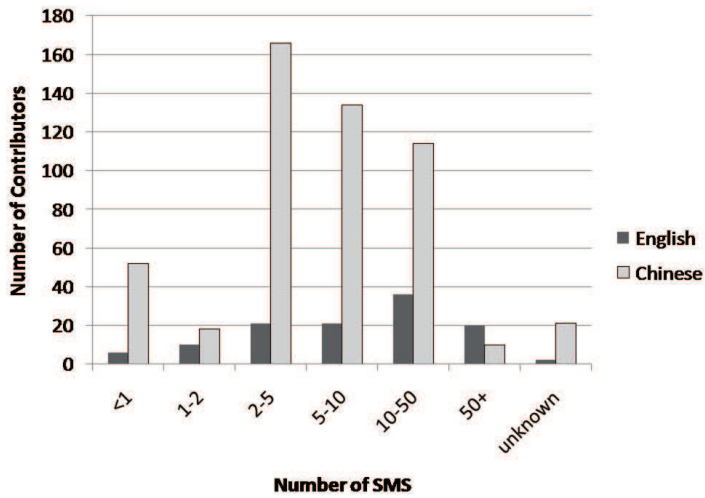}
     \caption{Contributors' Frequency of Sending SMS Daily}
    \label{fig:frequency}
\end{figure}

The distribution of 
the frequency of sending SMS is presented in Figure~\ref{fig:frequency}. We observe an interesting phenomenon that there is a gradually  increasing trend -- from 18.0\% (2--5), 18.1\% (5--10), to 31.0\% (10--50) -- for English contributors, while for Chinese contributors it decreases -- from 32.2\% (2--5), 26.0\% (5--10), to 22.1\% (10--50). Moreover, for the English portion, 17.2\% of contributor send more than 50 SMS everyday. 
We posit that the larger proportion of English contributors who SMS
more than 50 times a day may indicate that some use it to carry on
conversations (thus needing more messages) rather than for sending one-off messages.

\begin{figure}[h]
  \centering
    \includegraphics[width=0.8\textwidth]{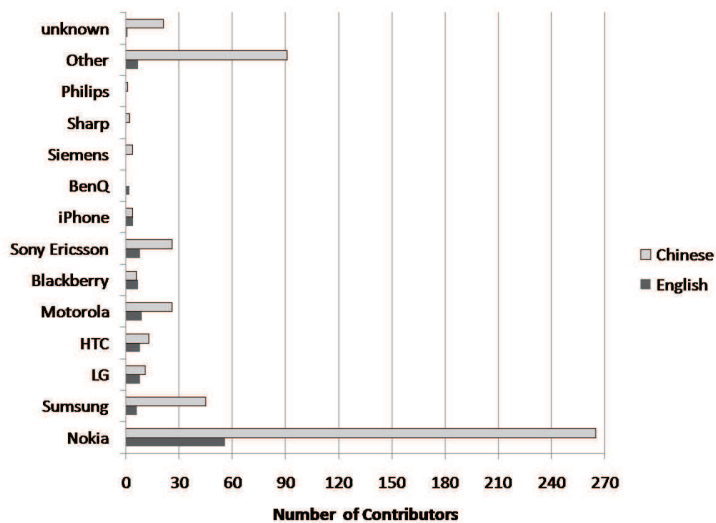}
     \caption{Contributors' Phone Brand}
    \label{fig:brand}
\end{figure}

Input method differences were also
revealed in our survey. Three common input methods, multi-tap, predictive and
full keyboard, account for  41.4\%, 30.2\% and 17.2\% of English SMS
contributors respectively. The remaining 10.3\% of contributors used other input
methods.  Chinese input methods also largely consist of three input methods,
pinyin, wubi and bihua; these account for 86.0\%, 3.9\% and 4.3\% of 
contributors' usage.  
As displayed in Figure~\ref{fig:brand}, the majority of contributors own a 
Nokia phone, which accounts for 48.3\% of English SMS contributors and 49.7\% of
Chinese SMS contributors. However, these figures cannot fairly present the
general popularity of phone brands, for the reason that we only provide links to
Nokia utility software for SMS Export collection method. In addition, 37.1\% of
English SMS contributors, and 44.9\% of Chinese SMS contributors own a
smartphone. This question was not included in the first version of survey used
in MTurk, which leads to smartphone information is unknown for 48.3\% of English
SMS contributors and 7.8\% of Chinese SMS contributors.

For more in-depth analysis beyond the scope of our article, we invite
you to visit the corpus website, where the full, per question answers
for our demographic data is available.

\subsection{Cost}

As MTurk, ShortTask and Zhubajie are crowdsourcing sites where participants
are motivated by profit, we compensated contributors monetarily.  The
same was true for the local collections that we ran.  For the calls for
participation via the Internet to linguistic, text corpus and phone
manufacturer's communities, we felt that their self-motivation to see the
corpus' success would be a strong motivator to participate.  Thus for these
communities, we did not provide any compensations.

Table~\ref{table:payment in mturk} shows the reward scheme used in our MTurk 
collection runs using the SMS Export and SMS Upload methodologies.
For example, 400 messages will be rewarded with \$4.50 in total with \$4.00 base
 payment and \$0.50 bonus. In ShortTask, we only published tasks using Web-based
Transcription method, with the reward of \$0.08 for 10 messages per submission.
Since Zhubajie was a new venue for crowdsourcing, we tuned our rewards scheme 
based on a pilot (as shown in Tables~\ref{table:payment1 in zhubajie} and
\ref{table:payment2 in zhubajie}).  Our pilot showed that workers were very
eager to participate, thus we decreased the reward amount in our subsequent
runs, which did not dampen workers' enthusiasm for completing the contribution
tasks.  For the four batches of English (2 batches) and Chinese (remaining two
batches) SMS Web-based Transcription, we instead aimed for breadth of
participants.  In each task, we recruited 20 workers with \$0.10 as the reward
for individual workers.  Finally, we also recruited contributors locally, whose
reward scheme is displayed in Table~\ref{table:payment in local}. 

Since the labor cost of collecting additional SMS over the first few is small 
(arguably even negligible in the Upload and Export methodologies), we
incentivize additional submitted messages with a bonus reward, which is based
on the number of messages that exceed the base range.  For all reward schemes, the
bonus amount diminishes with the amount of additional messages submitted, with a
maximum payment capped to allow us a measure of budgetary control.  Table~\ref{table:cost
comparison} shows the cost comparison between sources. We spent 92.30 US
dollars in MTurk, 2.94 US dollars in ShortTask, 868.50 Chinese yuan in Zhubajie
(all inclusive of commission fees) and 340.00 Singapore dollars for our
local collection. Standardizing all currency amounts to US dollars\footnote{On
23 October 2011, 1 SGD = 0.7848 USD, 1 CNY
= 0.1567 USD.} allows us to compute the cost per message for a clear
comparison.  In a nutshell,
local collection is the most expensive method, while
crowdsourcing remains an effective and economical method to collect our corpus
data.  We also note that Zhubajie is a little cheaper than MTurk, but that it
may only be applicable to certain languages (i.e., Chinese). Due to our
very limited usage of ShortTask, it doesn't make much sense to compare cost in
ShortTask with other sources.

\begin{table}[h]
\centering
 \caption{Reward Scheme in MTurk}
 \label{table:payment in mturk}
\begin{tabular}{l*{6}{c}r}
\hline
Total Number    		&  Base Reward  	& Bonus  \\ 
\hline
10--400 							& 0.10					& 1/100 Msg  		\\
401--1000            & 4.00 				& 1/200 Msg 	\\
$\geq1000$          & 7.00 				& 0 	\\
\hline
\end{tabular}
\end{table}

\begin{table}[h]
\centering
 \caption{Reward Scheme 1 in Zhubajie}
 \label{table:payment1 in zhubajie}
\begin{tabular}{l*{6}{c}r}
\hline
Total Number    		&  Base Reward  & Bonus  \\ 
\hline
10--100 							& 1.00					& 1/10 Msg  		\\
101--400            	& 10.00 				& 1/20 Msg 	\\
401--1000          	& 25.00				& 1/40 Msg 	\\
$\geq1000$          & 40.00				& 0	\\
\hline
\end{tabular}
\end{table}

\begin{table}[h]
\centering
 \caption{Reward Scheme 2 in Zhubajie}
 \label{table:payment2 in zhubajie}
\begin{tabular}{l*{6}{c}r}
\hline
Total Number    		&  Base Reward  & Bonus  \\ 
\hline
20--100 							& 1.00					& 1/20 Msg  		\\
101--400            	& 5.00 				& 1/25 Msg 	\\
401--1000          	& 17.00				& 1/40 Msg 	\\
$\geq1000$         	& 32.00				& 0	\\
\hline
\end{tabular}
\end{table}

\begin{table}[h]
\centering
 \caption{Reward Scheme for Local Collection}
 \label{table:payment in local}
\begin{tabular}{l*{6}{c}r}
\hline
Total Number    		&  Base Reward  & Bonus  \\ 
\hline
20--50 							& 2.00					& 1/10 Msg  		\\
51--200            	& 5.00 				& 1/30 Msg 	\\
201--600          		& 10.00				& 1/40 Msg 	\\
$\geq600$          	& 20.00				& 0	\\
\hline
\end{tabular}
\end{table}

\begin{table}[h]
\centering
 \caption{Cost Comparison}
 \label{table:cost comparison}
\begin{tabular}{l*{6}{c}r}
\hline
Source   	& Total Num    	& Total Cost  &		Cost Per Msg\\ 
\hline
MTurk 		& 11,330	& USD 92.30  		& USD 0.00815	\\
ShortTask	& 280		& USD 2.94		& USD 0.0105\\
Zhubajie        & 23,789	& CNY 868.50 		& CNY 0.0365 ($\sim$USD
0.0057) \\
Local         	& 20,245	& SGD 340.0 		& SGD 0.0168 ($\sim$USD
0.0132)	\\
\hline
\end{tabular}
\end{table}

\subsection{Towards a publicly-available live corpus}

Our corpus, consisting of both messages and associated user profiles, has been 
released publicly since February 2011\footnote{\url{http://wing.comp.nus.edu.sg/SMSCorpus}}. To achieve our goal of making an
accessible dataset, we have pursued an open license, public domain development
methodology that involved first the convincing and later the blessing of our
university's intellectual property department.  For the aim of making a general
purpose dataset, we have tried to incorporate a balanced approach for user
profiling; by requiring contributors to answer a set of demographics and
including them with the dataset.  So as to make the corpus as large as possible,
we incorporate all messages that we collected through all of the methodologies
used, although this means varying levels of quality among subportions of the corpus (e.g., some SMS may not have
an associated user profile).  

To make the corpus convenient for researchers to access, we also pioneer the 
distribution of the corpus both as an XML file as well as a database dump
in MySQL.  Potential SMS researchers or contributors can also browse and download
the corpus on the directly on the corpus website, and access dataset statistics,
all without the need to handle the raw corpus files or compute the statistics
themselves. Website access also shows the effect of the anonymization processes
discussed earlier, so that potential contributors can feel more secure knowing
how their messages will likely appear.

Our statistics help prospective users grasp a general understanding about the 
demographic and representativeness of our corpora. The corpus as well as
statistics will be updated on a monthly basis, since our collection is still in
progress. Moreover, the SMS can be directly browsed on our website, which
provides a convenient way to learn about our corpus without the need to process raw
files and helps potential contributors to understand our anonymization strategy
by viewing other's messages.  

\section{Discussion}
\label{sec:discussion}

We now comment on three open questions surrounding the crowdsourcing of our public corpus of SMS.
First, what do workers think about contributing SMS to a public corpus?
Second, how does the Chinese crowdsourcing site of Zhubajie compare with Amazon's Mechanical Turk? 
Third, as some crowdsourcing is motivated by altruism, how possible is it to collect SMS without offering any monetary reward?

\subsection{Reactions to our Collection Efforts}
The corpus was collected under our university's institutional review board (IRB) exemption policy and
important identifiers in the corpus have been replaced by placeholder
tokens for deidentification purposes.
However, our experience through on this project over the last year has shown us
that the privacy issues surrounding the collection of SMS is still
very much a concern.  Even among
our research group, members were largely unwilling to donate SMS even
with the safeguards in place.  This may have been partially due to
fact that the authors need to manually review the submitted SMS, and
that the review process may identify the contributor.  This fear was
further validated in our local collection drive, where potential
contributors worried that their friends may review the corpus and
identify their messages, especially through the mention of certain
names in SMS (Personal names are not replaced by any code as given in
\ref{table:replacement code}, as many personal names are also common
words as well).  

Privacy concerns were also paramount in our crowdsourcing work with
Amazon Mechanical Turk, and ultimately caused our account with MTurk to
be closed.  Amazon sent us several notices that our tasks violated
their terms of service.  Though private correspondence with Panagiotis
Ipeirotis, whose research involves detecting spam in MTurk, we
found out our tasks were routinely classified by turkers as a spam
or phishing task, despite our attempts to give credibility to the
project through the creation of the corpus webpage and browsing
interface\footnote{In fact, these were some of Ipeirotis' suggestions
  to ameliorate the problem, so credit is due to him.}.  Unfortunately,
even with repeated attempts to contact Amazon to clarify the nature of
our notice of breach of service, our MTurk account was suspended
without further detail from Amazon.

Similar concerns surfaced on our calls for SMS contribution in the Chinese realm.
Notably on the mobile phone forums that we posted our call for
participation, we encountered a few skeptic replies.  For these
forums, we had advertised our Android SMS uploader application, with
the appropriate links to an explanation of the project and our corpus'
web page.  Several posters voiced their concern that the software
might be malware looking to steal private information (especially
given the inclusive permissions set that our application needs access
to).  These were valid concerns as some previous mobile application
recommendations did turn out to be malware, so readers were being
cautious before installing any software.

\subsection{Zhubajie compared to MTurk}

Zhubajie is one of a growing number of middleware websites in China
offering crowdsourced labor.  Forecasted online transactions on the
site are expected to surpass 500 million CNY ($\sim$78 million USD) in
2011 alone\footnote{\url{http://economictimes.indiatimes.com/tech/internet/idg-backed-chinese-website-zhubajie-to-list-in-us-in-3-years/articleshow/9478731.cms/}}.

As discussed, we found Zhubajie to be a good platform to recruit
Chinese contributors. However, unlikely its western counterparts, Zhubajie, as well as other Chinese 
``witkey'' websites, has not been widely exploited for research data collection in
computer science community. Most exiting academic work focus on the business and
economic research, studying
the user behavior~\citep{Yang2008,DiPalantino2011,Sun2011}, analyzing the
participation structure~\citep{Yang2008-AAAI}, exploring the anti-cheating
mechanism~\citep{Piao2009}, and investigating the business
model~\citep{Zhang2011}. 
For these reasons, we feel it would be useful to give a more comprehensive
overview of Zhubajie, focusing on five aspects: its
conceptualization, characteristics of typical tasks, cost, completion
time and result quality.  We compare Zhubajie against the now familiar
Amazon Mechanical Turk (MTurk) when appropriate.

\begin{itemize}
\item {\bf Concepts.} While both Zhubajie and MTurk can be characterized as 
mechanized labor, to be more accurate, Zhubajie's form of crowdsourcing
originates from ``witkey'' -- an abbreviation of the phrase ``the key of
wisdom''.  The concept of Witkey was put forward by Liu in 2005, and later published in \citet{liu2007}, where its core concept
was to trade knowledge, experience and skill as merchandise.

\item {\bf Tasks.}
MTurk focuses on tasks that take a small or moderate amount of time to complete,
 while most tasks of Zhubajie require expertise and can take longer to complete.
 Designing logos, software development and revising resum\'{e} are typical
Zhubajie tasks.  Zhubajie also classifies tasks into a detailed hierarchical
classification system with major 18 top categories, and 6--22 secondary categories per top-level category.  It 
requires requesters to specify a secondary category for each task.  This is
unlike MTurk, which eschews task classification altogether.
Zhubajie's detailed browsable task hierarchy reflects its Chinese base, as the 
Chinese population often prefers selection and browsing over searching (as
browsing only requires clicking but searching requires inputting Chinese
characters, which is still somewhat difficult)\footnote{\url{http://www.smashingmagazine.com/2010/03/15/showcase-of-web-design-in-china-from-imitation-to-innovation-and-user-centered-design/}}.

This task classification leads to different service characteristics in Zhubajie 
and MTurk.  MTurk provides keyword search and 12 different sorting options for results display
(newest/oldest tasks, most/fewest available tasks, highest/lowest reward, etc.).
 The survey results of \citet{Chilton2010} shows that the newest and most
available tasks are the most popular sorting strategies employed by workers, and
that first two pages of result listings are most important.
\citet{IpeirotisMarket2010} pointed out that if a task is not executed quickly
enough, it will fall off these two preferred result listings and is likely to be
uncompleted and forgotten.  

This is in accord with our experience in trying to recruit workers. In Zhubajie,
even 10 days after publishing the task, we still received new  submissions from
workers. This is contrary to our experience with MTurk, where we did not receive many new submissions after
the first two days. Due to the detailed task hierarchy in Zhubajie, potential
workers can easily target specific tasks matching their expertise and interests,
ameliorating the recency-listing concerns prevalent in MTurk.  A few
outstanding workers, based on reputation and total earnings, are also featured
as top talents for each task category.  Requesters can invite and recruit
talents to fulfill the task.  These properties all help aid matching workers to
tasks in comparison to MTurk.

\item {\bf Cost.}
The demand of expertise in Zhubajie also impacts the payment distributions in 
two websites.  In MTurk, the lowest payment is just USD \$0.01 and 90\% of tasks
pay less than \$0.10~\citep{IpeirotisMarket2010}. Compared to the tiny rewards
offered in MTurk, the rewards in Zhubajie are significantly higher, with about
\$0.15 (CNY 1.00) as the lowest reward and about \$182 (CNY 1181) as the average
reward for the year 2010.  Also, though both services make profit by collecting
commission fees, they differ as to whom the commission is charged from: MTurk
charges the requester 10\%, but Zhubajie charges the worker 20\% commission.
Furthermore, In Zhubajie, task rewards come in two flavors: they can be set by 
requesters or they can be bid on by workers.  In Zhubajie, these are referred to
as {\it contract tasks} and {\it contest tasks}, respectively. In this sense,
our task -- and MTurk tasks in general -- are thus contract tasks. For our SMS
collection thus far, Zhubajie has turned out to be more economical by 30.1\%; we
spent \$0.0057 and \$0.00815 per message in Zhubajie and MTurk, respectively.

\item{\bf Completion Time.}
Here we look at the task completion time with respect to collection
methodology.  With the SMS Upload method, it took 2 full days to
receive 3 English submissions via MTurk; and worse, there were no
submissions at all from Zhubajie.  This may be explained by the
current low popularity of Android smartphones among Chinese SMS contributors.  In contrast, under the SMS Export
collection method, we received 16 Chinese submissions from Zhubajie in 40 days, and
27 English submissions from MTurk in 50 days. The collection in this method was
slow in both platforms.

Web-based Transcription offers the most telling demographic difference.
Our MTurk tasks took 2 days to complete, collecting 40 valid English
submissions and 20 days for 20 valid Chinese submissions (each submission
having 2 or 5 individual SMSes). In contrast, the same Chinese SMS task, when
published to Zhubajie, usually took less than 30 minutes to complete to collect
for 20 submissions. We ascribe the quick completion in Zhubajie to two reasons. First, 
Zhubajie has a specific task category for SMS tasks -- the {\it creative greetings}
category.  This category typically asks workers to compose a creative SMS and
send it to bless a designated recipient (i.e., write a poem to wish someone to
get well soon), as it is uplifting in China to receive lots of blessing from the
general public.  It is also a relatively popular category among workers for its
short completion time and low requirements for expertise.  Second, compared to
other tasks in the creative blessing category, our task is easier, faster and
more profitable.  Other tasks require workers to design or craft a creative
blessing and send it to an actual recipient which incurs cost, but the payment
is usually identical to ours.

\item {\bf Quality} has emerged as a key concern with crowdsourcing, and it is 
clear that this is a concern for our task as well.  MTurk employs several
strategies to help requesters control for quality: a requester can require
certain qualifications based on the worker's location and approval rate,
publishing a qualification test prior to real task and blocking certain poor
workers from a task.  To attract the maximal number of potential contributors,
we did not set any qualifications in MTurk.  Zhubajie does not provide built-in
quality control system. Tasks, when completed in either MTurk or Zhubajie, can be rejected by the requester if it does not meet 
their criteria for a successful task.  Table~\ref{table:approval comparison}
shows our approval rate of completed tasks for each collection method in the two
crowdsourcing websites.

As we have previously described the problems with Web-based Transcription (in 
that contributors can enter anything they want, including SMS copied from other
SMS sites on the web), we expected this poorly-performing methodology to have
the highest rejection rate.  Surprising, in fact, it was quite the opposite:
Web-based Transcription tasks enjoyed a higher approval rate than the other
methods, across both sites. We believe the difference in financial incentives of
the methodologies explains this.  While the payment for Web-based Transcription
was only US \$0.10 in MTurk and CNY 1.00 in Zhubajie, the payment of the other
two methods can be as high as US \$7.00 in MTurk and CNY 40.00 in Zhubajie.
Intrigued by the high reward, some workers attempted to cheat on these
higher-yield methods. This validates findings by~\citet{Mason2009}, who states
that increased financial compensation may not improve task quality, and  sometimes may 
even result in poorer quality.

SMS Upload approval rates were also relatively better than those for SMS Export.
 Workers using SMS Upload needed to type a unique code generated by the mobile
application, which may discourage errant workers from cheating since they would have not known how to generate the correct code without doing the task through the application.  In contrast, the SMS Export method allowed contributors to upload
any files (even those not containing SMS at all), making it easier for potential
cheaters to try their luck.

Finally, we judged the overall quality of work done by Zhubajie workers to be 
much higher than that of MTurk. We attribute this to the open worker reputation
system of Zhubajie. In MTurk, the worker's approval rate is the sole figure to
judge whether a worker's work is good or bad. Zhubajie stores comments on
workers as well as calculating a positive comment rate (similar to MTurk's
approval rate) and a reputation rank which is based on earned income. In some
cases, if many workers compete for one task, the requester can pick over the
potential, qualified workers based on these positive comment and reputation
rank.  Finally, Zhubajie's administrators will warn serious cheaters and even
lock their account.  MTurk metes out no official punishment for cheaters (unlike
our experience for requesters), and requesters have to manually blacklist poor
workers in their tasks.

\begin{table}[h]
\centering
 \caption{Approval Rate Comparison}
 \label{table:approval comparison}
\begin{tabular}{l*{6}{c}r}
\hline
Collection Method   	& MTurk   		& Zhubajie \\ 
\hline
Web-based Transcription & 62.50\%		& 85.03\% \\
SMS Export            	& 16.58\%		& 57.14\% \\
SMS Upload         	& 42.86\%		& No submissions \\
\hline
\end{tabular}
\end{table}

\end{itemize}

\subsection{Altruism as a Possible Motivator}

Mechanized labor is just one possible form of crowdsourcing that can result in workers performing a task.  There have been a number of surveys on crowdsourcing, including a recent survey on finding optimal the method for crowdsourcing corpora and annotations \cite{wang2011}.  Are other, non-profit oriented approaches feasible for collecting sensitive data?  Could altruistic motivational factors work?

To answer this question, we emailed calls for participation to the natural language and corpora community\footnote{via the Corpora List, corpus4u forum, and the (Chinese) 52nlp blog.} for voluntary, non-compensated contributions of SMS.  This was a part of our methodology from the beginning as described earlier.  Unfortunately, the results were not promising. As shown in Table~\ref{table:sms and sources} of Section~\ref{sec:statistics}, we received only 5 anonymous contributions, totalling 149 English and 236 Chinese SMS, respectively by these methods\footnote{From additional personal contacts, we obtained an additional 319 English and 1476 and Chinese SMS respectively.}.  Compared to the rest of the for-reward collection methods, this method was a failure, and we do not recommend this method for collection in its current guise.

Our findings are contrary to the sms4science project, which succeeded in 
gathering a large number of messages through pure voluntary contribution. Though
a small portion of contributors were randomly selected by lottery to win 
prizes\footnote{\url{http://www.smspourlascience.be/index.php?page=14}}, we still
deem their collection method as a purely voluntary contribution, as there is no monetary
compensation. However, we note two key differences between our call and theirs.
The sms4science project obtained support from phone operators, to make it free
for potential contributors to forward their SMS to a service number. This lowers
the difficulty of contributing messages as no software installation or tedious
export is necessary (but note it does destroy some message metadata that we can
collect through our other methods).
Probably more important was that the sms4science project conducted large-scale
publicity; its call for participation was broadcast in the main national media
including press, radio, television~\citep{Fairon2006,durscheid2011}. For
example, the Belgium project was reported  in five newspapers and six websites
within two weeks
\footnote{\url{http://www.smspourlascience.be/index.php?page=16}}. 
For our project, due to the limited publicity
vehicles and technical constraints, attracting people to contribute SMS only for
the sake of science was difficult. 

Our appeal to the research community did not yield many SMS for the corpus, but did give  us further convictions that we were performing a necessary and useful task.  Several researchers supported our project by writing words of encouragement and sharing their personal difficulties with gathering SMS for research.  

\section{Conclusion}
\label{sec:conclusion}

In order to enlarge our 2004 SMS corpus and keep up with the current technology 
trends, we resurrected our SMS collection project in October 2010
as a live corpus for both English and Mandarin Chinese SMS.  Our aim in
this revised collection are threefold: to make the corpus 1) representative for
general studies; 2) accurate with fewer transcription errors; 3) released to the
general domain, copyright-free for unlimited use; 4) useful to serve as a reference dataset. To achieve these four goals,
we adopt crowdsourcing strategies to recruit contributors from a wide spectrum
of sources, using a battery of methodologies to collect the SMS.  
As SMS often contains sensitive personal data, privacy and anonymization
issues have been paramount and have influenced the resulting design of the
collection methods and the corpus data itself.

We are very encouraged by the results so far.  As of the October 2011 version of
 the corpus, we have collected 28,724 English SMS and 29,100 Chinese SMS, with a
cost of 497 U.S. dollar equivalent and approximately 300 human hours of time
(inclusive of the Android app implementation, website creation and
update). 
As the project is a live corpus project, all of these figures are growing as the
collection is a continuing effort.  To the best of our knowledge, our corpus is
the largest English and Chinese SMS corpus in the public domain.  We hope our
corpus will address the lack of publicly-available SMS corpora, and enable
comparative SMS related studies.

A novel aspect of our collection is the implementation of mobile phone
applications for collection.  We adapted an SMS backup software to also serve as
a platform for contributing SMS.  It is the first application for such purpose
and is easily adapted for other SMS collection purposes.  We report on the use
of the first use of Chinese crowdsourcing (also known as ``Witkey'') for
collecting corpora and have discussed the significant differences between
Chinese and traditional crowdsourcing in the English-speaking world, as embodied
by Amazon's Mechanical Turk.  Finally, we explored the possibility of calling
for SMS contribution without compensation, but found that altruistic motivation is
not sufficient for collecting such data.  Rather, our lessons learned indicate
that large-scale publicity is the key to success.

We continue to enlarge our SMS collection, as part of our interpretation of what
 a live corpus project means.  Given the importance of SMS in carrying personal
communication in our society, and the low-cost methods we have found to collect
message contents (suitably scrubbed) and demographic data, we are encouraged to
continue to fund this work internally, to encompass more languages and a wider
population of users.  We also plan to explore other SMS collection methods, such
as an iOS (i.e., iPhone, iPad) application, and benchmark their efficacy against the
methods we have analyzed in this article.

As for downstream use, with further funding, we may annotate the corpus (either 
automatically or manually) for part-of-speech, translation to other languages,
or other semantic markup.  The resulting corpus may then be used in other
natural language studies and applications (e.g., machine translation).  Other
downstream projects that we know of may also make their annotations and
additional collection of SMS available as collaborative or sister projects to
our NUS SMS corpus.


\section{Data}
\label{sec:data}
The corpus described in this paper is publicly available at our corpus website (\url{http://wing.comp.nus.edu.sg/SMSCorpus}).

\begin{acknowledgements}
We would like to thank many of our colleagues
who have made valuable suggestions on the SMS collection,
including Jesse Prabawa Gozali, Ziheng Lin, Jun-Ping Ng,  Yee Fan Tan, Aobo Wang and Jin Zhao.
\end{acknowledgements}


\bibliographystyle{chicago}      

\end{document}